\newif\ifusenix
\newif\ifacm
\newif\ifmcom
\newif\ifieee

\usenixfalse
\acmfalse
\mcomfalse
\ieeetrue

\ifieee
    \documentclass[letterpaper, 10 pt, conference]{ieeeconf}  

    \IEEEoverridecommandlockouts                              
\fi

\ifusenix
	\documentclass[letterpaper,twocolumn,10pt]{article}
	\usepackage{usenix2019_v3}

	\usepackage{times}
\fi

\ifacm
  \documentclass[sigconf,10pt]{acmart}
  \renewcommand\footnotetextcopyrightpermission[1]{} 

\fi

\ifmcom
  \documentclass{sig-alternate-10pt}
\fi

\usepackage{xcolor}
\usepackage{placeins}
\usepackage{amsfonts}
\usepackage{balance}
\PassOptionsToPackage{hyphens}{url}\usepackage{hyperref}
\usepackage{color}
\usepackage{graphics}
\usepackage{graphicx}
\usepackage{url}
\usepackage{listings}
\usepackage{multicol}
\usepackage{multirow}
\usepackage[scaled]{helvet}
\usepackage{rotating}
\usepackage{xspace}
\usepackage[ruled,vlined]{algorithm2e}
\usepackage{comment}
\usepackage{amsmath}
\usepackage{mathrsfs}
\usepackage{cancel}
\usepackage{textgreek}
\usepackage{authblk}
\usepackage[font=small,labelfont=bf]{caption}
\usepackage[skip=0pt]{subcaption}
\usepackage{microtype}
\usepackage{multirow}

\usepackage[capitalise]{cleveref}
\crefname{section}{Sec.}{Sec.} 
\crefname{algocf}{Alg.}{Algs.}
\usepackage[explicit,noindentafter]{titlesec}

\titlespacing\section{1pt}{1pt plus 2pt minus 2pt}{2pt plus 2pt minus 2pt}
\titlespacing\subsection{1pt}{3pt plus 2pt minus 2pt}{1pt plus 2pt minus 2pt}
\titlespacing\subsubsection{1pt}{3pt plus 2pt minus 2pt}{1pt plus 2pt minus 2pt}

\DeclareMathOperator*{\argmin}{argmin}
\DeclareMathOperator*{\argmax}{argmax}

\newcommand{\name}{ViWiD\xspace}
\renewcommand{\iota}{\textsl{j}\xspace}

\title{
\name: Leveraging WiFi for Robust and Resource-Efficient SLAM
}

\author{Aditya Arun$^{1}$, William Hunter$^{1}$, Roshan Ayyalasomayajula$^{1}$, and Dinesh Bharadia$^{1}$
\thanks{$^{1}$ UC San Diego, CA, USA
        {\tt\small \{aarun,wshunter\}@ucsd.edu}\\
        {\tt\small\{roshana,dineshb\}@ucsd.edu}}
}
        
\ifacm

\begin{abstract}
    Recent interest towards autonomous navigation and exploration robots for Indoor applications has spurred research into deploying Simultaneous Localization and Mapping (SLAM) systems for Indoor applications.
    While most of these SLAM systems use Visual and LiDAR sensors in tandem with a Odometer sensor, these Odometer sensors drift over time.
    This leads to the Visual SLAM systems to use compute and memory intensive algorithms to correct for this drift and perform SLAM.
    In this paper, we present \name, which integrates WiFi and Visual (camera) sensors in a dual-layered system.
    This Dual-layered approach provides independent tasks to each layer making \name memory and compute efficient while achieving comparable performance to state-of-the-art Visual SLAM.
    We demonstrate \name's performance on a dataset we have collected, spanning $25 \times 30$ m, and robustly tested by driving the robot a cumulative distance of over $1000$m.
    Wherein we show that we consume 5$\times$ less compute and 3$\times$ less memory on average compared to state-of-the-art Visual SLAM system.
\end{abstract}
\fi

\begin{document}

\maketitle

\ifieee
    \thispagestyle{empty}
    \pagestyle{empty}

\begin{abstract}
    Recent interest towards autonomous navigation and exploration robots for indoor applications has spurred research into indoor Simultaneous Localization and Mapping (SLAM) robot systems.
    While most of these SLAM systems use Visual and LiDAR sensors in tandem with an odometry sensor, these odometry sensors drift over time.
    To combat this drift, Visual SLAM systems deploy compute and memory intensive search algorithms to detect `Loop Closures', which make the trajectory estimate globally consistent.
    To circumvent these resource (compute and memory) intensive algorithms, we present \name, which integrates WiFi and Visual sensors in a dual-layered system.
    This dual-layered approach separates the tasks of local and global trajectory estimation making \name resource efficient while achieving on-par or better performance to state-of-the-art Visual SLAM.
    We demonstrate \name's performance on four datasets, covering over $1500$ m of traversed path and show 4.3$\times$ and 4$\times$ reduction in compute and memory consumption respectively compared to state-of-the-art Visual and Lidar SLAM systems with on par SLAM performance.
    
    \textit{Keywords} -- Sensor Fusion; SLAM; Localization

\end{abstract}
\fi

\section{Introduction}\label{sec:intro}

Diverse indoor applications are increasingly interested in deploying autonomous indoor robots.
These robots are typically equipped with Simultaneous Localization and Mapping (SLAM) frameworks to enable real-time navigation and to generate a human-readable map of the robot's environment.
To enable these applications, most state-of-the-art SLAM systems~\cite{Labb2018RTABMapAA, rosinol2021kimera, cartographer, mur2015orb} use visual (monocular or RGB-D cameras) and/or LiDAR sensors in tandem with odometry measurements reported from IMUs (inertial measurement unit) or wheel-encoders to locate themselves and map the environment. 
Despite this fusion, error in the predicted trajectory increases over time due to accumulation of sensor errors. Fortunately, these drifts can be corrected with `loop closures'~\cite{arshad2021role} which correlate the current observation with a dictionary of past observations, ensuring self-consistency of the robot's estimated trajectory on a global scale.

Unfortunately, these much-needed loop-closures are also the weakest links in SLAM systems, as they increase the memory requirements, are compute intensive, and are not robust~\cite{arshad2021role}.
In particular, false-positive loop closures, common in monotonous or visually dynamic environments, are highly detrimental to robot pose predictions. Furthermore, as we scale to larger spaces, performing loop closures demands the storage of an ever-larger dictionary of unique observations, demanding higher memory usage. Consequently, the feature matching needed to discover loop closures in large spaces requires extensive search, leading to high compute requirements. Thus, for real-time SLAM systems to be both accurate and scalable, it is imperative they are resource-efficient. Hence, we explore if these resource consuming loop closures can be entirely removed while maintaining SLAM accuracy.

To develop a SLAM system without loop closures, we need to find an alternative method to re-identify previously visited spaces and make the current pose estimate consistent with past estimates in that space.
Incorporating static and \textit{uniquely identifiable} landmarks in the environment aids in re-identifying previously visited spaces. Furthermore, \textit{accurately mapping} their poses allows the robot to leverage these landmarks to anchor its estimate to the environment and circumvent loop closures. Accordingly, various static landmarks which are identifiable by cameras~\cite{olson2011apriltag} or LiDARs~\cite{huang2021lidartag} have been deployed where loop closures are insufficient to correct for drifts.  
Unfortunately, these visual landmarks can fail in situations of blockage or dynamic lighting conditions. Their deployment further scales poorly to large environments. 

\noindent\textbf{Goal:} Clearly, to deliver a resource-efficient and real-time SLAM system, we need to incorporate `uniquely identifiable' and `mappable' landmarks in our environments. These landmarks need to be robust in dynamic environments and easily scalable to large spaces.  

To achieve this goal, we demonstrate that WiFi access points are a robust replacement to these visual landmarks and can aid in removing `loop-closures'. We develop \name which integrates them with visual sensors for accurate and resource-efficient SLAM without relying on loop closures. \name is also readily scalable, as WiFi access points are already ubiquitously deployed in indoor environments making them a natural choice for a landmark.

\subsection{Literature Review}
Most existing Visual and LiDAR SLAM algorithms try to resolve memory issues in loop-closure detection and identification by: (a) optimizing key-frame and key-point detection~\cite{mur2015orb} to reduce the number of key-frames stored, or (b) using a smart representation, like bag-of-words~\cite{rosinol2021kimera} for efficient storage and retrieval. While these solutions reduce the amount of memory required per step, the underlying problem of the memory consumption being linear in the length of the robot's trajectory remains. This implies that even the most efficient representations will eventually run out of memory given a large or complex enough environment.

Past work~\cite{arun2022p2slam} demonstrated accurate WiFi based SLAM, which has no need for loop closures as its WiFi measurements exist in a globally consistent reference frame, even if the positions of the anchors are unknown beforehand. However, it performed only an a-posteriori fusion of wheel odometry and WiFi measurements and cannot perform online robot operation.
There are also works like WSR~\cite{jadhav2020wsr}, which demonstrate robot reconnaissance using WiFi and not global indoor navigation. Most other existing WiFi-based SLAM systems~\cite{hashemifar2019augmenting, liu2019collaborative,huang2011efficient,ferris2007wifi} depend only on WiFi RSSI readings which are unreliable in dynamic indoor scenarios~\cite{ma2019wifi} and require a-priori fingerprinting data for navigation.

Other RF-sensor technologies that are used for localization which are shown to extend to robotic navigation include UWB~\cite{nguyen2021viral}, BLE~\cite{sato2019rapid, jadidi2018radio}, RFID~\cite{ma2017drone}, or backscatter devices~\cite{zhang2020robot}. However, these RF-sensors are \textbf{(a)} not as ubiquitously deployed limiting wider adoption or \textbf{(b)} have shorter ranges compared to WiFi limiting their scalability to large spaces. 

\subsection{Challenges}
In this paper, we present \name, "\underline{Vi}sual \underline{Wi}Fi \underline{D}ual-graph", a dual-layered real-time SLAM system which overcomes the need of loop-closure detection making it resource efficient. To build \name, we surmount the following challenges:

\noindent\textbf{(a) Tradeoff between WiFi measurement accuracy and compute:} We have seen from existing works~\cite{arun2022p2slam, jadhav2020wsr, zhang2020robot} that bearing measurements extracted from the WiFi Channel-state estimate (CSI) are viable for WiFi based SLAM algorithms. These systems use \textit{super-resolution} algorithms like MUSIC and SpotFi~\cite{spotfi} to obtain low-noise bearing measurements needed for online operation. However, this accuracy comes at the cost of computation, and are not suitable for realt-time deployment on devices with low compute capabilities.

\noindent \textbf{(b) Infeasible sensor fusion via a single factor graph:} Most sensor-fusion techniques rely on incorporating measurements within factor graphs~\cite{gtsam} to optimally solve for the SLAM problem. Unfortunately, WiFi is subject to outlier measurements due to transient reflections, so immediate incorporation of every measurement can hurt the local state estimate. Hence, a simple integration of WiFi landmarks into the same framework as a Visual/LiDAR sensor~\cite{rosinol2021kimera} can lead to a poor estimate of the recent robot poses due to local inconsistency in WiFi measurements.

\noindent\textbf{(c) Initialization of Unknown WiFi anchors:} As new landmarks, visual or WiFi-based, are discovered, they need to be accurately placed in the map to ensure stable convergence of the factor graph. 
This task is easy in the case of visual markers'~\cite{olson2011apriltag, huang2021lidartag} owing to the pixel-scale accuracy of cameras and LiDARs. Unfortunately, the poorer resolution offered by WiFi measurements makes initialization of the AP's position estimate in the environment nontrivial. A poor initialization can lead to the non-optimal state estimate or worse, an indeterminacy in the solution. 

\subsection{\name's Contributions}
To overcome these challenges, \name deploys a bifurcated design: (a) a third-party visual-inertial odometry (VIO) module that can provide accurate and real-time odometry measurements at a local scale, and (b) and a WiFi sensor module that can plugs into an existing VIO system and performs online correction of global drifts.
Using these insights, we make the following contributions to enable accurate, real-time indoor navigation for robots.

\noindent\textbf{(a) Accurate and compute-efficient WiFi Measurements:} 
\name breaks away from the accuracy-compute tradeoff by designing a PCA-based WiFi Bearing (PCAB) estimation algorithm. By adequately combining WiFi measurements over time to suppress noise, PCAB circumvents compute intensive super resolution algorithms whilst delivering accurate and real-time bearing estimates.

\noindent\textbf{(b) Extensible dual graph optimization:} To make best use of the WiFi and visual sensors, \name proposes to construct a Visual-WiFi Dual graph system. We utilize local odometry measurements (without global loop closure detection) extracted via inertial sensors and visual feature tracking. These local odometry measurements are then fused with WiFi measurements to track the WiFi landmarks in the environment by our WiFi graph. This dual-graph approach further allows robots to plug-and-play \name into the existing Visual/LiDAR SLAM systems whilst reducing their compute and memory consumption. 

\noindent\textbf{(c) Smart initialization of WiFi landmarks:} Finally, to dynamically map these WiFi landmarks when they are first observed, the robot tracks the strength of the WiFi signal from the access point (AP). Next, we initialize the APs location close to the robot's current location when we see an inflection point in the change of signal strength measured. This allows us to initialize the AP close to its true location, which improves the convergence of the factor graph. 

To verify our claims, we have deployed \name on a ground robot TurtleBot2 platform that is equipped with a Hokuyo LiDAR and Intel Realsense D455 RGB-D camera with a built-in IMU for deploying cartographer~\cite{cartographer} and Kimera~\cite{rosinol2021kimera} respectively. We also equip it with a 4 antenna WiFi radio~\cite{nexmon:project}. We deploy the robot in one large environment to collect data for demonstrating \name's compatibility with Kimera's VIO outputs (with loop-closures turned off), \textit{in addition} to three open-sourced datasets~\cite{arun2022p2slam} that we use to demonstrate \name's deployability with LiDAR-inertial odometry (LIO) from Cartographer. Across these deployments the robot traverses for an overall time of $108$ minutes and a distance of $1625$ m. 
We show that \name achieves a median translation error of $70.8$ cm and a median orientation error of $2.6^{\circ}$, on par with the state-of-the-art Kimera~\cite{rosinol2021kimera} and Cartographer~\cite{cartographer}. While achieving a similar navigation accuracy: (a) \name only needs a total of $0.72$ GB for a $25$ minute run, wherein Kimera needs $2.82$ GB, (b) \name utilizes on average $0.72$ fraction of single core of CPU whereas Cartographer utilizes over $3.2$ cores of the CPU on average. Thus \name demonstrates accurate, low-compute and low-memory SLAM.  

\section{\name's Dual layered Design}\label{sec:design}

\name seeks to deploy a SLAM system which can be compute and memory efficient. However, most current visual-based SLAM systems rely on resource-intensive loop closures to correct for inevitable drifts in robot's trajectory predictions and ensure consistency. Fortunately, a solution to circumvent these loop closure operations exists -- deploy \textit{identifiable} and \textit{mappable} landmarks in the environment to anchor the robot to it's environment and help it provide accurate trajectory estimates. But most of the current visual markers fail in dynamic environments or are tedious to deploy. To improve the robustness of these unique landmarks and hence guarantee a consistent trajectory, \name leverages WiFi access points in the environments as the much-needed landmarks. WiFi access points are uniquely identifiable from the hardware MAC address and can be mapped in the environment as shown by recent work~\cite{arun2022p2slam}, thus meeting the two criteria for a landmark. Moreover, unlike aforementioned visual landmarks, WiFi is ubiquitously deployed in indoor environments, and it is unaffected by visual occlusions or dynamic environments.

However, we are presented with three important challenges. \textit{First}, for the access point to be `mapped' into the environment, we require consistent and error-free WiFi measurements from the WiFi signals received at the robot from these access points (\cref{sec:design-bearing}). \textit{Second}, these WiFi measurements need to be appropriately optimized with visual and odometry sensors to provide a globally consistent trajectory (\cref{sec:design-graph1}). And \textit{third}, the optimizer needs to be well initialized to ensure timely convergence so that a real-time trajectory may be furnished (\cref{sec:design-init}). In the following sections we will seek to surmount these challenges to deliver \name, an accurate and resource efficient SLAM system. 



\subsection{Providing accurate and real-time WiFi measurements}\label{sec:design-bearing}

\begin{figure}
    \centering
    \includegraphics[width=0.8\linewidth]{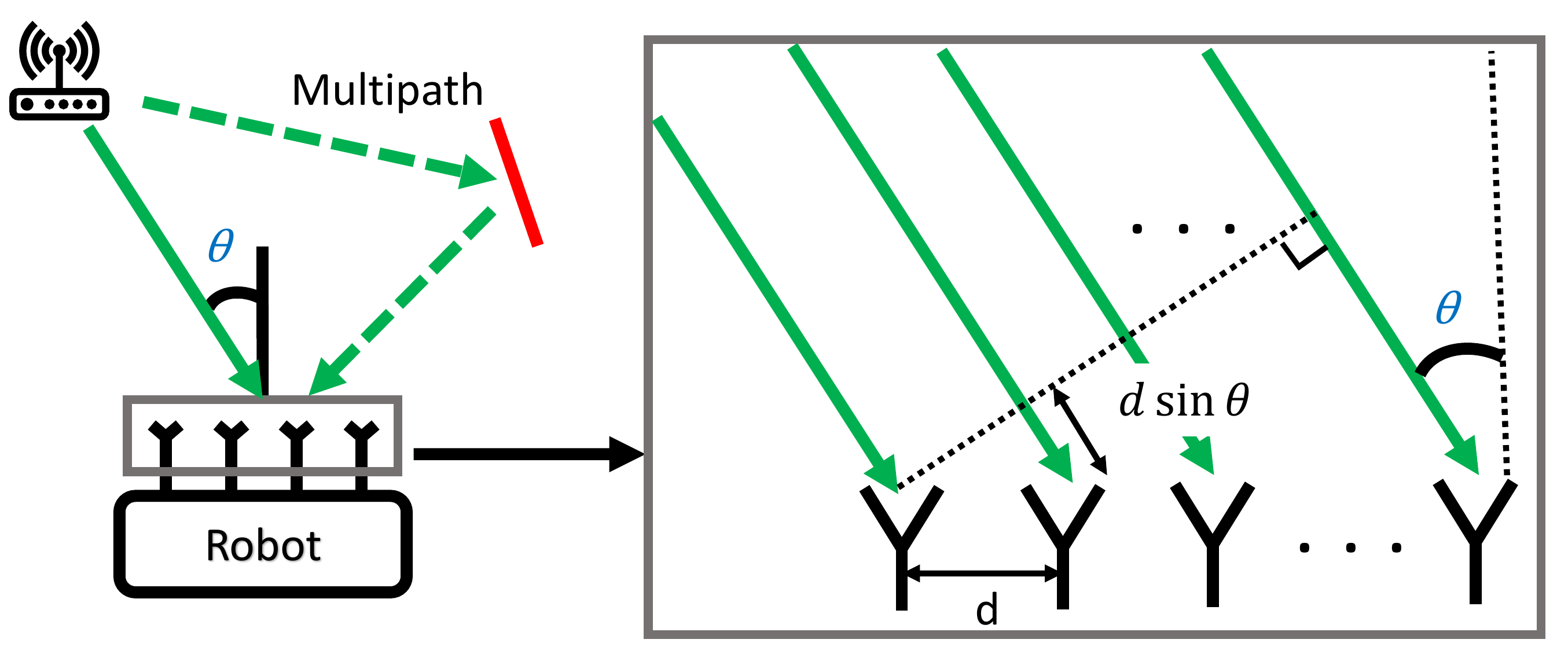}
    \caption{Shows WiFi bearings ($\theta$) measured using a linear antenna array with antenna separation $d$. The additional distance $d\sin(\theta)$ travelled by the signals can be exploited to estimate the bearing}
    \label{fig:wifi-meas}
\end{figure}

Recent research into decimeter accurate WiFi localization~\cite{ma2019wifi,spotfi} and accurate WiFi SLAM~\cite{arun2022p2slam,jadhav2020wsr,zhang2020robot} has shown that bearings are a viable WiFi-based measurement for indoor localization and tracking.
Inspired from these works, we use the bearing measured from the incoming signals at the robot as our WiFi measurements (as shown by the solid line in Figure~\ref{fig:wifi-meas}) to help the robot `map' the access points in the environment. 
Specifically, these bearing measurements at the robot provide the direction of the WiFi AP's in the robot's local frame along the azimuth plane. In fact, this is similar pixel coordinates of a corner feature in an image frame which provide its bearing in both azimuth and elevation. 

These WiFi-bearings can be estimated using the Channel State Information, CSI, measured on reception of a WiFi signal~\cite{arun2022p2slam,spotfi}. However there are \textit{three} challenges which need to be surmounted for accurate bearing estimates, \textbf{(a)} multiple reflected copies of the same WiFi signal arrive at the robot and corrupt the bearing measurements; \textbf{(b)} concrete walls or metal reflectors can block the direct path signals and create inconsistent bearing estimates; \textbf{(c)} linear antenna array geometries commonly deployed in \cite{arun2022p2slam, spotfi} can reduce the diversity of bearings which can be measured definitively and in turn reduce the number of measurements which can be incorporated into the system. Let's tackle the first challenge by modelling the WiFi channel in the environment. 

\noindent\textbf{Accurate and Real-time bearing measurements:} The primary reason for inaccuracies in the bearing measurements for indoors is multipath~\cite{spotfi,ayyalasomayajula2020deep,ma2019wifi}. 
A WiFi signal, with wavelength $\lambda$, is broadcast, and multiple reflections of the signal (\textit{multipath}) along with the direct path, impinge at the receiver as shown in Figure~\ref{fig:wifi-meas} (left). Clearly, the `direct-path' (solid-line) is the only path helpful in estimating the bearing ($\theta$) to the source and the rest of the `reflected-paths' (dotted lines) are the cause for erroneous bearing estimates. 

To understand these effects, we provide a simple mathematical model. The receiver measures, at time $t$, a complex-valued channel state information (CSI) describing the phase delay and attenuation across each of the $M$ receiver antennas and $N$ orthogonal frequencies~\cite{spotfi} as,
\begin{align}
    X^{m, n}_t &=  a_{m, n} e^{j \phi_m (\theta)} e^{-j (2\pi f_n \tau + \psi)}; \quad X_t \in \mathbb{C}^{M \times N} \nonumber \\ 
    \phi_m(\theta) &= -\frac{2\pi}{\lambda} m d \sin(\theta) \label{eq:angle-phases}
\end{align}
where, $a_{m, n}$ is the attenuation; $d$ is the antenna separation for the linear antenna array; $f_n$ is the orthogonal frequency; $\tau$ is the time-of-travel of the signal, which is often corrupted by the random $\psi$ phase offset due to lack of transmitter-receiver clock synchronization; and $\phi_m (\theta)$ is the additional phase accumulated at the $m^\mathrm{th}$ antenna (Figure~\ref{fig:wifi-meas}) due to the additional distance travelled by the signal. 
The various reflected paths impinging on this linear array will add to each antenna and frequency component in a similar manner. 

To identify the direct signal's path among the multiple reflected paths, past Wi-Fi systems\cite{ma2019wifi} have used algorithms that `super-resolve' the measured signal. By using information across these $N$ different frequency bins, they measure the relative time offset between different signal paths. This allows identification of the direct path, as it must have traveled the least distance of all paths and thus arrives before the other reflections. However, the addition of this extra dimension of time-offset adds computation overhead, making them unsuitable for resource-efficient SLAM algorithms. 


So, in \name, we reduce the dimensionality of our problem and yet reliably segregate the direct path signals from the clutter of the reflected paths. Specifically, from the channel model described in Spotfi~\cite{spotfi}, the largest eigenvector ($U_t \in \mathbb{C}^{M}$) of $X_tX_t^H \in \mathbb{C}^{M \times M}$ provides the largest contribution to the channel measured across the $M$ receive antennas. However, this largest component could potentially be corrupted by multipath. To remove the effect of multipath, we make a simple observation -- reflected paths are susceptible to small changes in the robot's position as opposed to the direct path which will arrive at a consistent bearing. Hence we can effectively `average-out' the effects of multipath from our bearing estimation if we can combine our measurements across time (over multiple packets). We can effectively do this by finding the largest eigenvector over most recent $T$ measurements across the past $.5$ seconds:
$$
U_t = \boldsymbol{\lambda}(\sum_{i= t-T}^t X_{i}{X}_{i}^H),
$$
Where $\boldsymbol{\lambda}(\cdot)$ extracts the largest eigenvector, which is trivial to compute for our $M \times  M$ autocorrelation matrix. From here, our direct path signal $U_t$ can be mapped to a bearing by a coarse search over the space of possible bearings,  
\begin{align}
    \theta^* = \argmax_{\theta \in [-90^\circ, 90^\circ]} \sum_{i=1}^M \phi_m(\theta) U_t(m) \label{eq:bearing-search}
\end{align}
This method, dubbed Principle Component Analysis based Bearing (PCAB), allows us to achieve similar bearing estimation performance to the standard 2D-FFT algorithm~\cite{arun2022p2slam} and Spotfi~\cite{spotfi} at just a fraction of the compute. 

\noindent\textbf{RSSI Filtering:} But we have made an implicit assumption in PCAB that a direct path signal from an access point is present. However, in cases where a large reflector blocks the direct path, no information can be obtained about the direct path's bearing, and hence can lead to inconsistent bearing measurements. These non-line-of-sight (NLOS) scenarios are common and need to be handled adequately to avoid instability of the factor graph. We observe that in these situations, overall received signal power (RSSI) is typically very low, as the majority of the signal has been blocked. Hence, we reject all measurements with RSSI below $-65$dBm, which we empirically observe filters out the vast majority of these obstructed packets. Thus, \name provides real-time, unambiguous and accurate bearing measurements to be fed into the WiFi factor graph, described in ~\cref{sec:design-graph1}.

\begin{figure}[t]
    \centering
    \includegraphics[width=0.42\textwidth]{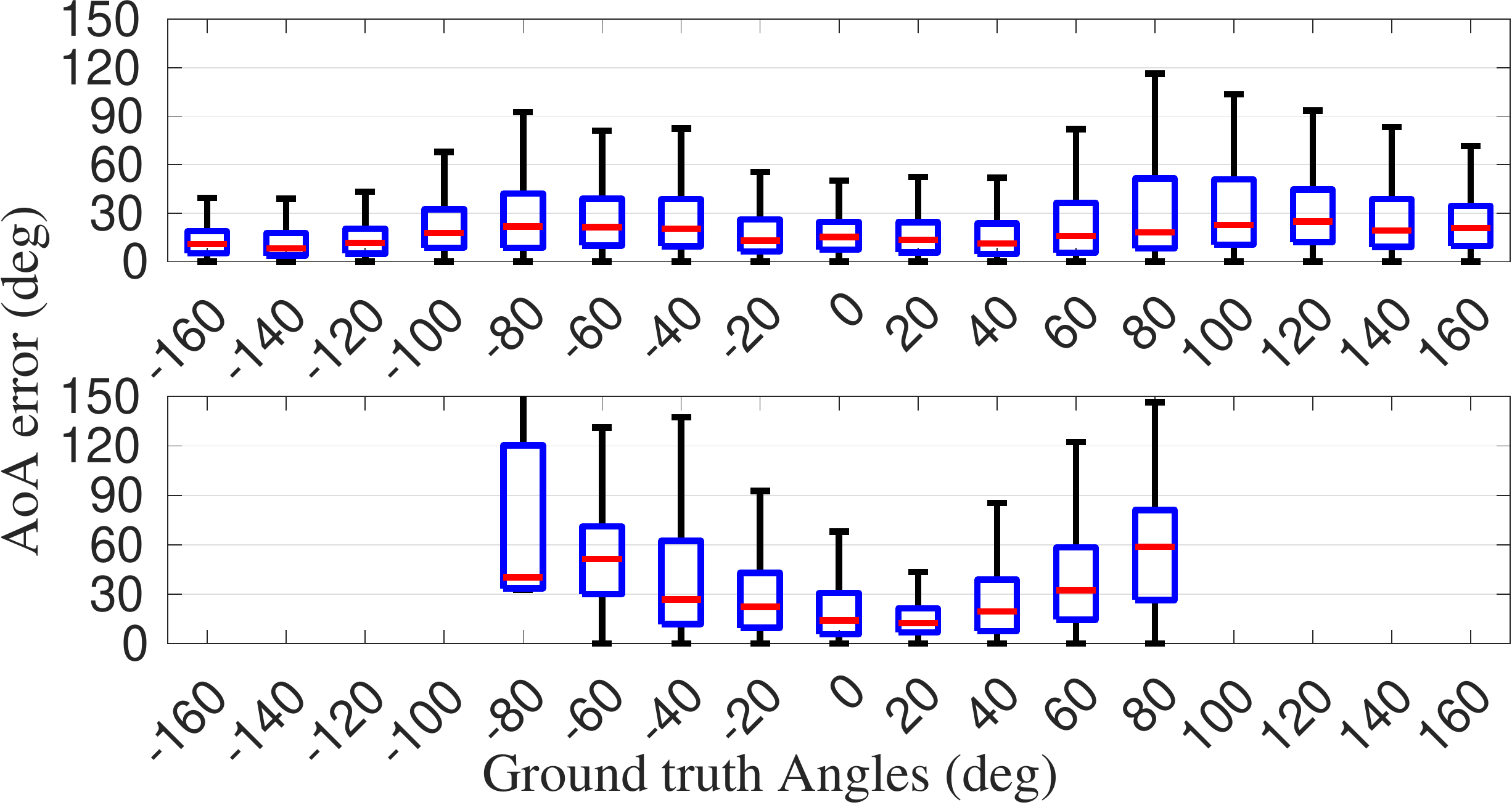}
    \caption{Bearing errors are accumulated in steps of $10^\circ$ for the the square antenna array (\textit{Top}) and linear array (\textit{Bottom}).}
    \label{fig:aoa-array}
\end{figure}

\noindent\textbf{Unambiguous Measurements:} Finally, to enable reliable `mapping' of our WiFi landmarks, we would like to measure bearings from the largest range of angles. Unfortunately, the uniform linear arrays used by existing bearing estimation algorithms~\cite{spotfi, arun2022p2slam} are vulnerable to aliasing - they can only measure bearings in a $180^\circ$ range(i.e. the top half plane in Figure~\ref{fig:wifi-meas}) at a time, as there is no distinction between signals coming from opposite sides of the array. This is further seen in the search space of bearing angles as in Eq~\ref{eq:bearing-search}. Furthermore, as shown in Figure~\ref{fig:aoa-array}(bottom), these arrays have poorer accuracy when WiFi signals arrive nearly parallel to the array (near $\pm 90$). To maximize the range of measured angles, we resolve this ambiguity by adopting a square antenna array, which does not suffer from aliasing. 
This can be further seen from Figure~\ref{fig:aoa-array}(top) -- \name finds a good trade-off between resolution and aliasing, and is able to resolve angles from $-160^{\circ}$ to $160^{\circ}$. Note however we cannot extend the range to the entire $360^\circ$ due to ambiguity present for WiFi signals arriving from behind the robot. 
Transitioning to a square array however changes the differential phases measured (Eq~\ref{eq:angle-phases}) as, 
\begin{align}
    \phi_m(\theta) = -\frac{2\pi}{\lambda} (X_m \cos(\theta) + Y_m \sin(\theta)),
\end{align}
the relative position of antenna $m$ is $(X_m, Y_m)$ with respect to the first antenna.
    
Finally note that these CSI measurements need to be sanitized beforehand the of random phase offsets ($\psi$) and a one-time calibration needs to be applied as further explained in Spotfi~\cite{spotfi}. Hence, by averaging-out multipath over multiple consecutive packets, rejecting NLOS measurements via RSSI filtering and incorporating a square antenna array at the robot, we have effectively furnished low-compute and accurate bearings over a $320^\circ$ range of angles to reliably `map' our WiFi landmarks. Next, we tackle how to incorporate these WiFi measurements into our optimizing backend to correct for trajectory drift.

\begin{figure}[t]
    \centering
    \includegraphics[width=\linewidth]{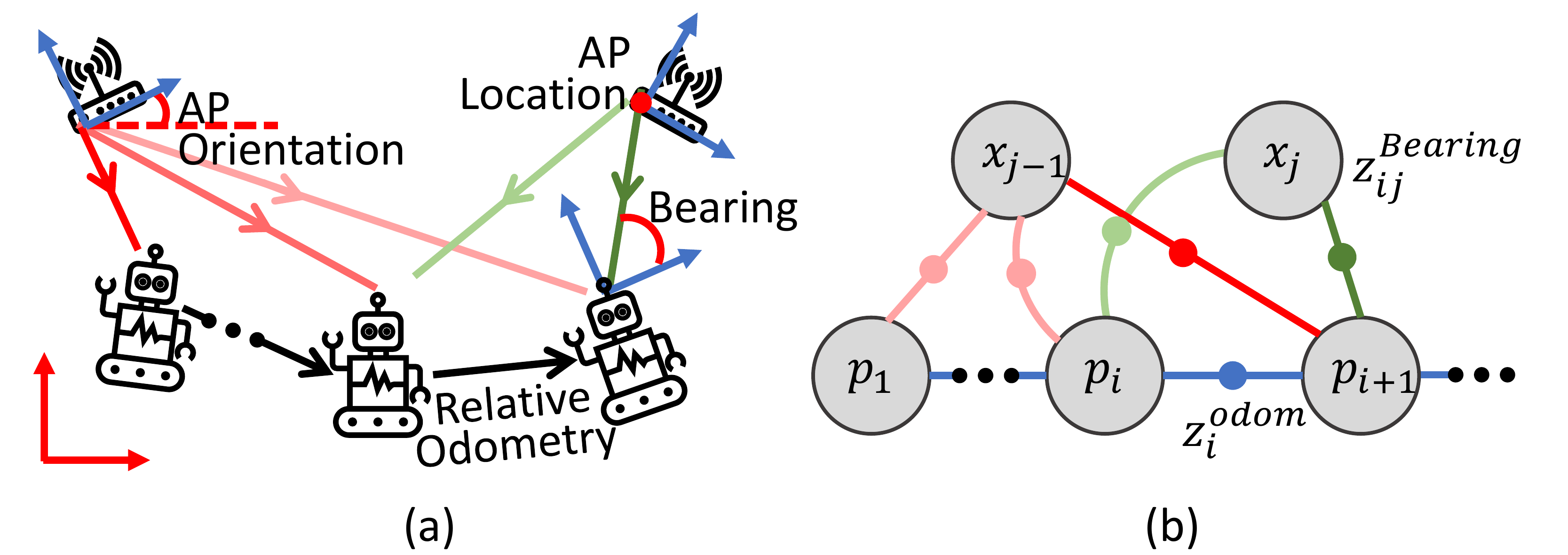}
    \caption{\name's WiFi Graph (a) Shows the various measurements that are made across robot poses and AP poses. (b) Shows the details of how these measurements are laid as factors in the implementation of the WiFi Factor graph}
    \label{fig:wifi-factor}
\end{figure}



\subsection{Building and Optimizing the WiFi-Graph}\label{sec:design-graph1}



Given these WiFi-bearing measurements, the first idea would be to integrate them within the factor graph of an available Visual Factor graph~\cite{rosinol2021kimera}. Unfortunately, discovery and addition of new AP's and global drift corrections introduce brief periods of instability (order of few seconds) to the robot's trajectory estimates. These instabilities can introduce large computation overheads as it may demand corrections to the tracked visual landmarks as well. To isolate these periods of instability, we propose a dual graph approach, where the drift-corrected poses from the WiFi graph can be utilized for globally consistent mapping. But unlike prior work~\cite{arun2022p2slam, ferris2007wifi}, we do not use end-end optimization and instead opt to use incremental smoothing and mapping (iSAM~\cite{gtsam}) to provide real-time pose estimates, which demands a more accurate and realtime bearing estimation as achieved in Sec.~\ref{sec:design-bearing}. 

To build the WiFi graph, consider the state space at time $t$, $S_t$. It is a set of robot poses and access point locations over $t$ time steps in our graph, $S_t = \{\Vec{p}_i \; \forall i \in [1, t]\} \cup \{\Vec{x}_j \; \forall j \in [1, N]\}$, with the robot pose, $\Vec{p}_i \in SE(3)$ and the $N$ access points positions observed till time $t$, $x_i \in \mathbb{R}^3$. 
We can the define odometry measurements (from VIO/LIO or wheel encoders, as shown in Fig~\ref{fig:wifi-factor}) between poses $\Vec{p}_i$ and $\Vec{p}_{i+1}$ at two consecutive time steps as:
\begin{align*}
    \hat{z}_i^{odom}(\Vec{p}_i, \Vec{p}_{i+1}) = \begin{bmatrix} 
                    R(p^q_i)^{-1} (p^t_{i+1} - p^t_{i})^T\\
                    (p^q_{i+1} \ominus p^q_i)_{[1:3]}
                \end{bmatrix} \nonumber 
\end{align*}
where, $R(\cdot) \in SO(3)$ is the rotation matrix corresponding to the given quaternion, $\ominus$ is the relative difference between two quaternions, and $[1:3]$ chooses only the first three elements of the quaternion.
Similarly, the bearing factors from  AP $j$ at robot position $x_i$ is $\hat{z}^{Bearing}_{ij}(p_i, x_j) \in \mathbb{R}^2$:
\begin{align}\label{eq:ping-pred}
    &\hat{z}^{Bearing}_{ij}(p_i, x_j) = \mathrm{Local}(\mathrm{TransformTo}(x_j, p_i), p_i) \\
                                &= \mathrm{Local}([\cos(\phi)\cos(\theta), \cos(\phi)\sin(\theta), \sin(\phi)]^T, p_i) \nonumber
\end{align}
where, \textit{TransformTo}$(\cdot)$ transforms the coordinates of the AP $x_j$ to the coordinate system provided by the robot at $p_i$, in which the AP subtends an elevation angle of $\phi$ and an azimuth angle of $\theta$. \textit{Local}$(\cdot)$ projects this bearing measurement to the tangent plane defined by the current pose of the robot, $p_i$.

Having defined the measurement models, we can estimate the optimized robot poses $S^{opt}_t$ for time $t$ by minimizing the total error between our predictions and actual measurements,
\begin{align}\label{eq:graph-optim-head}
    S^{opt}_t = \argmin_{S_t} & \sum_{i \leq t} \sum_{j \leq N}  \rho\left(\big(e^{bearing}_{ij}\big)^T \Sigma_\mathrm{bearing}^{-1} e^{bearing}_{ij};c\right) \nonumber\\
         & + \sum_{i\in I_{sub}}  \big(e^{odom}_i\big)^T \Sigma_\mathrm{odom}^{-1} e^{odom}_i
\end{align}
where $e^{bearing}_{ij} = z^{Bearing}_{ij} - \hat{z}^{Bearing}(p_i, x_j)$ is the bearing factor's error between robot and AP poses $i$ and $j$; $e_i^{odom}(p_i, p_{i+1}) = z_i^{odom} - \hat{z}_i^{odom}(p_i, p_{i+1})$ is the odom factor's error; $\rho$ is the Huber cost function with parameter $c$~\cite{zhang1997parameter}. $\Sigma_\mathrm{odom} \in \mathbb{R}^{6 \times 6}$ and $\Sigma_\mathrm{bearing} \in \mathbb{R}^{2 \times 2}$ are diagonal covariance matrices for odometry and bearing measurements respectively. Further note that the bearing measured in Sec.~\ref{sec:design-bearing} measured $\theta$, the azimuth angle in the robot's local frame and we assume that the elevation, $\phi$, of the incoming signal is $0$, hence $z^{Bearing}_{ij} = [\cos(\theta) \sin(\theta)]$. We can assume the elevation angle is $0$ despite AP's placed at differing heights as it has little affect to the azimuth bearing estimation~\cite{arun2022p2slam}.

\subsection{Initialization of Factors}\label{sec:design-init}
Next, we initialize each of the $t$ robot positions and $N$ AP positions for the optimizer. 
Similar to prior works~\cite{arun2022p2slam}, the robot positions are initialized using the relative odometry measurements. These poses have accumulated drift over time which we seek to correct.
A naive initialization for the AP's would be to place them at the origin and allow the optimizer to place them appropriately in the environment~\cite{arun2022p2slam}. Unfortunately, the iSAM optimizer, given the limited number of measurements it has seen until time $t$, can fail to converge with this naive initialization. 


To solve for this in-determinancy due to poor initialization of the AP's position estimate, we draw from the intuition of WiFi signal propagation characteristics.
WiFi's received signal signal strength indicator (RSSI), has been studied extensively in the past for localization.
While RSSI measurements are unreliable to perform accurate localization, they can still provide a general sense of proximity.
Thus, we can identify the robot's position $p^T_{ap-j} \in \mathbb{R}^3$ at which the $j^{th}$ AP's RSSI measurement has an maxima inflection point among all current robot's poses $p_i \forall i \in [1,t]$, with the intuition that this is where the robot passed closest to the AP.
We can then initialize the $j^{th}$ AP's pose $x_j$ as $x_j = p^T_{ap-j} + \Delta$, where $\Delta \in \mathbb{R}^3$ is a small ($\leq 0.1 m$) random perturbation. This random perturbation is added to avoid in-determinancy with the optimizer. Finally, with the real-time, compute-efficient WiFi-bearing in hand, along with a reliable way to integrate these measurements with local odometry measurements, we can proceed with the graph optimization iteratively with each time-step. In the following section we will provide specific implementation details.


\section{Implementation}\label{sec:implementation}

\noindent{\textbf{Hardware: }}\label{sec:implement-soft} We implement \name on the Turtlebot 2 platform. For WiFi CSI data collection, we attach an off-the-shelf WiFi radio~\cite{nexmon:project} to the Turtlebot. Then we place a few of the similar APs near the ceiling at a density of roughly one every ten meters. We transmit on channel 42 of 5GHz WiFi. We then use an open-sourced toolbox~\cite{nexmon:project} to collect channel state information (CSI) data from all the APs at the WiFi radio on the robot.
Our robot is also equipped with a Hokuyo UTM-30LX LiDAR and a Intel D455 RGBD camera with an IMU.

\noindent\textbf{Software:}
Given this setup, we compare our system against the realtime performance of two systems (a) Kimera~\cite{rosinol2021kimera}, a state-of-the-art VIO that uses the Intel D455 + IMU for its SLAM, and (b) Cartographer~\cite{cartographer}, a state-of-the-art LIO that uses the wheel encoders and the Hokuyo LiDAR for its SLAM.
The Turtlebot is controlled with a laptop running Robot Operating System (ROS-Kinetic), which manages all sensor input. WiFi CSI measurements are also integrated into ROS, and \name is implemented as a C++ application to ensure realtime operation and integrated within ROS as a ROS-node, allowing for easy integration with other robotics systems.
The Wi-Fi factor graph is implemented in GTSAM~\cite{gtsam}, and we utilize the open-sourced library to implement Kimera and Cartographer. \name's codebase will be open-sourced upon the paper's acceptance.

\begin{figure*}[t]
    \begin{minipage}{0.24\linewidth}
        \centering
        \includegraphics[width=\linewidth]{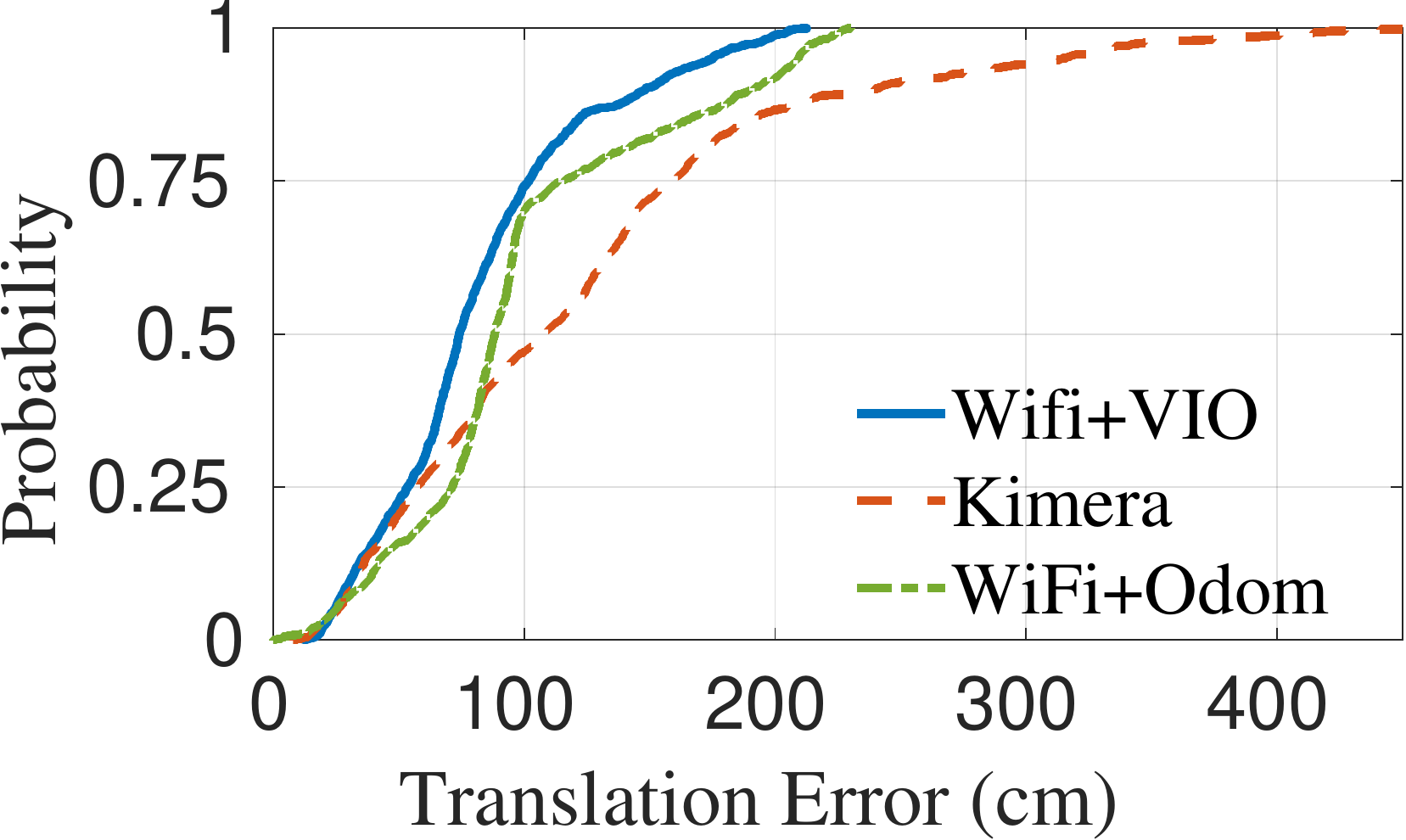}
        \subcaption{}
    \end{minipage}
    \begin{minipage}{0.24\linewidth}
        \centering
        \includegraphics[width=\linewidth]{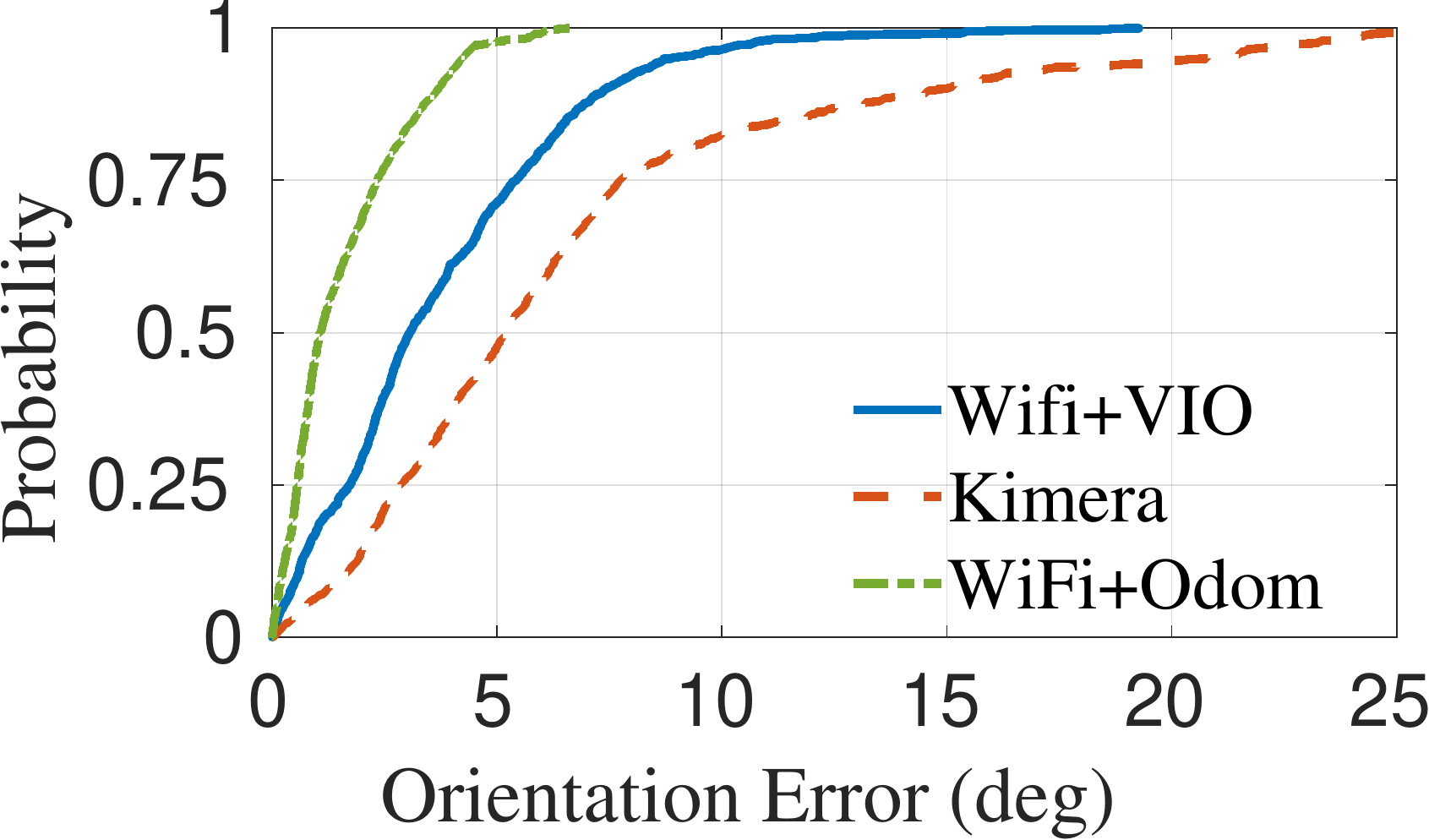}    
        \subcaption{}
    \end{minipage}
    \begin{minipage}{0.24\linewidth}
        \centering
        \includegraphics[width=\linewidth]{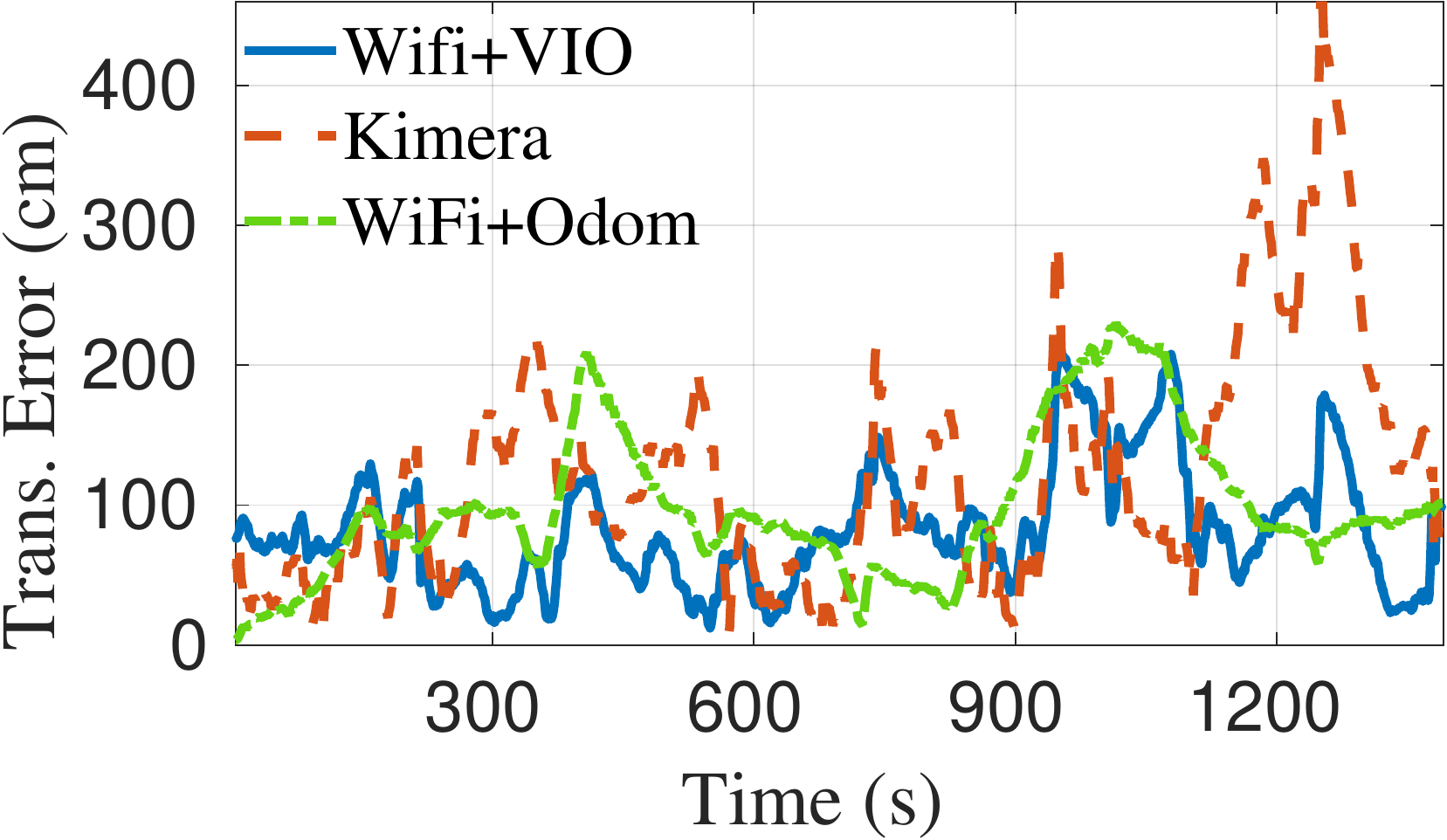}    
        \subcaption{}
    \end{minipage}
    \begin{minipage}{0.24\linewidth}
        \centering
        \includegraphics[width=\linewidth]{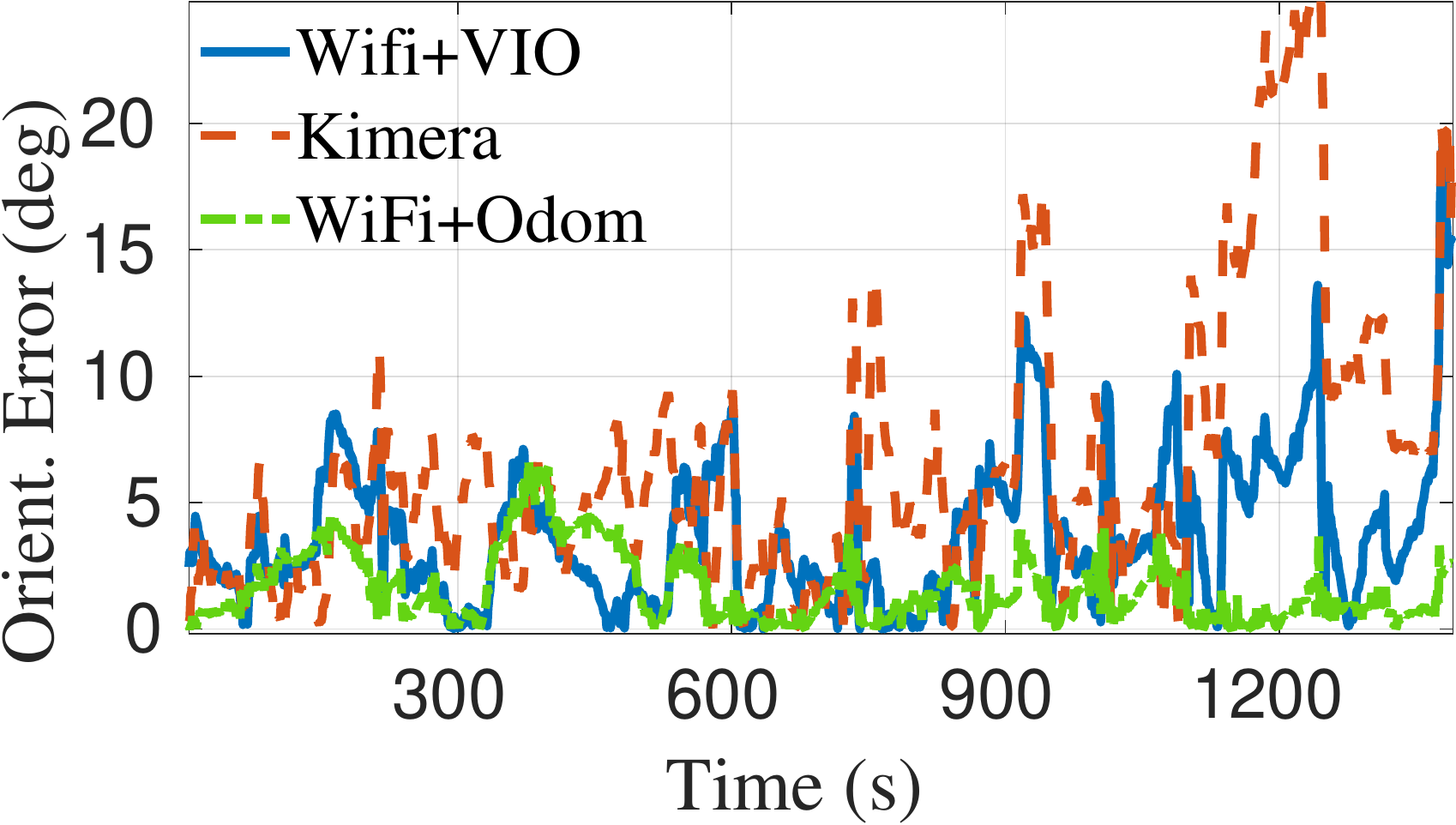}    
        \subcaption{}
    \end{minipage}
    
    \caption{\textbf{End-to-end evaluation}:  \textbf{(a, b)} Translation and orientation errors comparing Kimera with loop closure against \name's predictions. \name uses odometry from Kimera without loop closures and just wheel odometry . \textbf{(c, d)} Time series of translation and orientation error comparing Kimera with loop closure against \name's predictions. \name uses odometry from Kimera without loop closures.}
    \label{fig:end-end1}
\end{figure*}

\section{Results}\label{sec:eval}

\begin{figure}[t]
    \centering
    \begin{minipage}[t]{0.18\textwidth}
        \includegraphics[width=\textwidth]{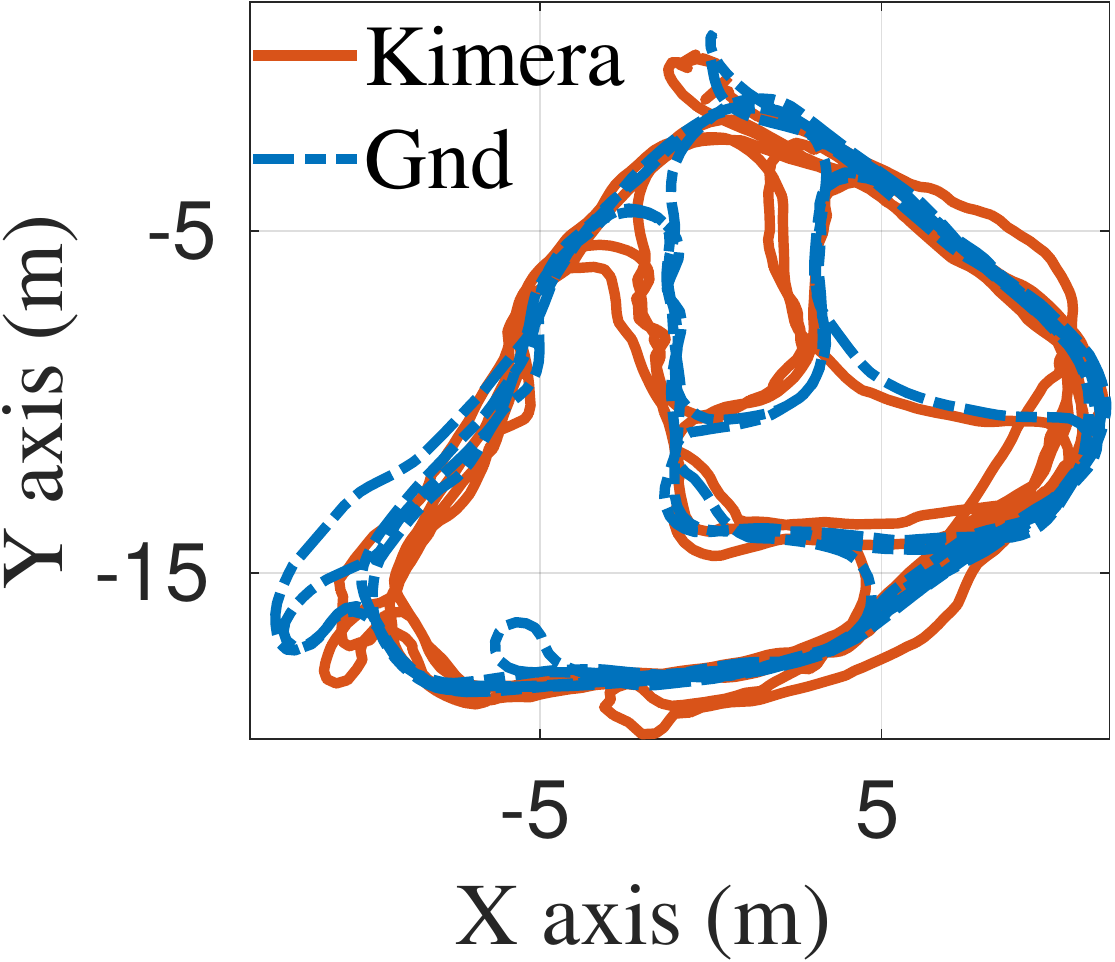}
    \end{minipage}
    \; \;
    \begin{minipage}[t]{0.18\textwidth}
        \includegraphics[width=\textwidth]{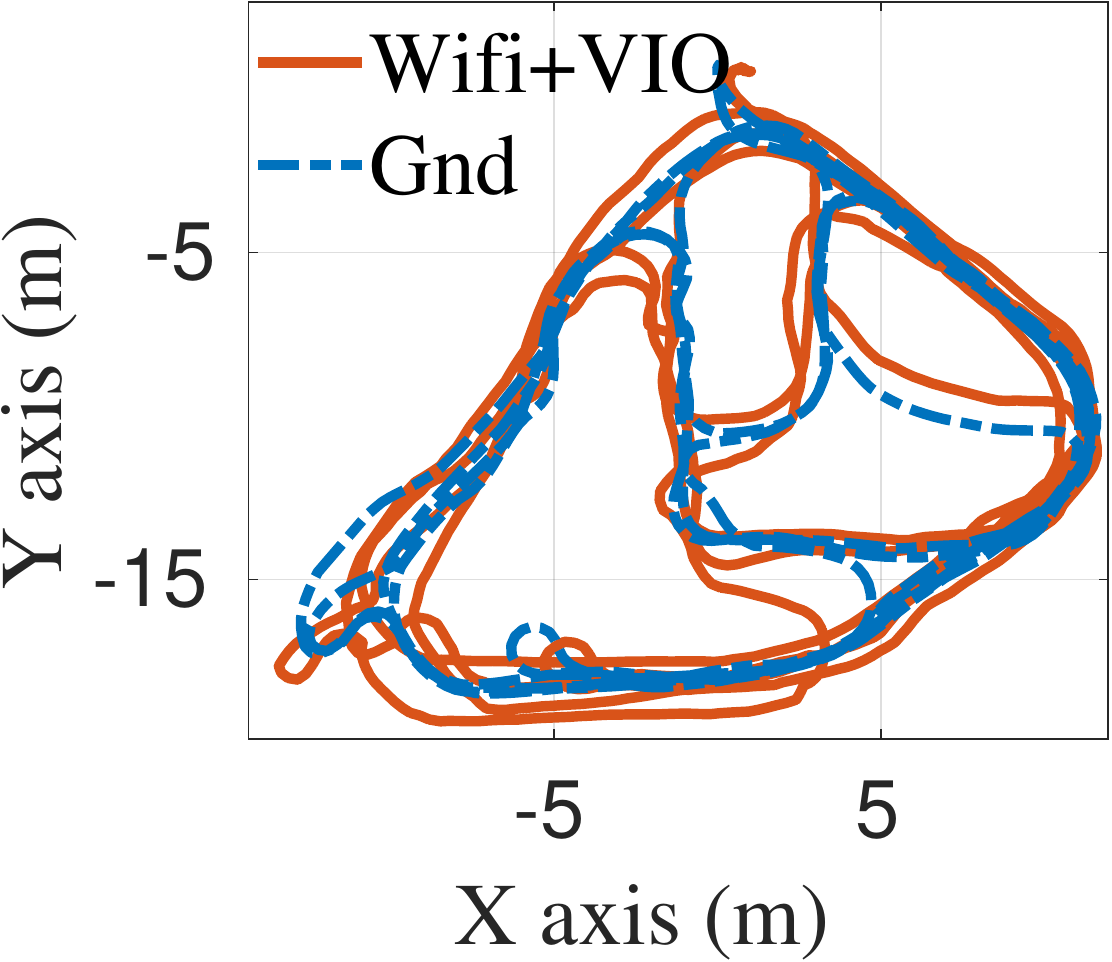}
    \end{minipage}
    \caption{\textbf{Trajectories:} A top-down view showing the Trajectories estimated by \textbf{(a)} Kimera with loop closures and ground truth. \textbf{(b)} \name using odometry predictions from Kimera without loop closures (\textit{WiFi + VIO}) and ground truth. }
    \label{fig:end-end2}
\end{figure}

Next, we demonstrate \name's performance in 4 datasets and compare it with state-of-the-art SLAM systems Kimera~\cite{rosinol2021kimera} (VIO) and Cartographer~\cite{cartographer} (LIO). Across these datasets, we compare three aspects of each algorithm's performance: 

\noindent\textbf{(a)} \textbf{3-DoF Navigation Accuracy}: XY euclidean translation error and absolute orientation error, 

\noindent\textbf{(b)} \textbf{Memory Consumption:}\footnote{\url{https://github.com/alspitz/cpu_monitor}} The total memory consumed for a run of the dataset, and also the rate of memory consumption, to understand scalability to larger indoor spaces, and 

\noindent\textbf{(c)} \textbf{Compute Cost:} As the max and average number of cores required by the algorithm to perform online SLAM.

\noindent\textbf{Datasets:} To demonstrate our system against Kimera (which requires a stereo-camera), we deploy a robot in a $20 \times 25$ m environment with 3 WiFi APs, with the robot traversing a total distance of $403$ m over a duration of $23$ minutes (called \name DS). We also use the open-sourced datasets~\cite{arun2022p2slam} that we call DS 1/2/3. But these 3 datasets do not have stereo-camera data, and use a linear antenna array on the robot, for which the aliasing is resolved using ground truth information. Through these 4 datasets, we have robustly tested \name across 3 distinct environments, 4 different and realistic access point placements, and traversed a cumulative distance of $1625$ m with a total travel time of $108$ minutes. To obtain ground truth labels, we allow Cartographer to run offline with very extensive search parameters, which converges to a near ground truth trajectory as characterized in DLoc~\cite{ayyalasomayajula2020deep}. Finally we intend to open source this dataset for the benefit of the larger research community up on the paper's acceptance. 

\noindent\textbf{Baselines:} We compare \name's performance with (a) Kimera in the \name DS, (b) Cartographer in all the datasets. Kimera and Cartographer have their loop-closures fine-tuned to provide the best realtime performance. 


\name's modular dual-graph design enables the WiFi-graph to receive odometry measurements from different SLAM systems. Thus, we give \name the following sources of odometry:
\textbf{(a)} \textbf{WiFi+VIO}: Online Kimera without the Loop Closure detection node,
\textbf{(b)} \textbf{WiFi+LIO}: Online Cartographer without global scan matching, and
\textbf{(c)} \textbf{WiFi+Odom}: The built-in wheel encoder on the TurtleBot.
Further note that we disabled visualization on Kimera and Cartographer to only account for the memory and computation by the optimization backend and keyframe storage.

We compare Kimera with WiFi+VIO and Cartographer with WiFi+LIO and provide an in-depth analysis of results and plots for Kimera~\cref{sec:eval-kimera} and summarize the results for Cartographer in a table in ~\cref{sec:eval-cart} due to space constraints.
Finally, in ~\cref{sec:eval-micro} we demonstrate the fine-tuning and trade-offs for both Kimera and Cartographer.

\subsection{End-to-end evaluation (\name with Kimera):}\label{sec:eval-kimera}

First, we show how \name provides a resource-efficient alternative to Kimera's loop closure detection module with on-par or better navigation accuracy.

\begin{figure}[t]
    \centering
    \begin{minipage}[t]{0.42\textwidth}
        \includegraphics[width=\textwidth]{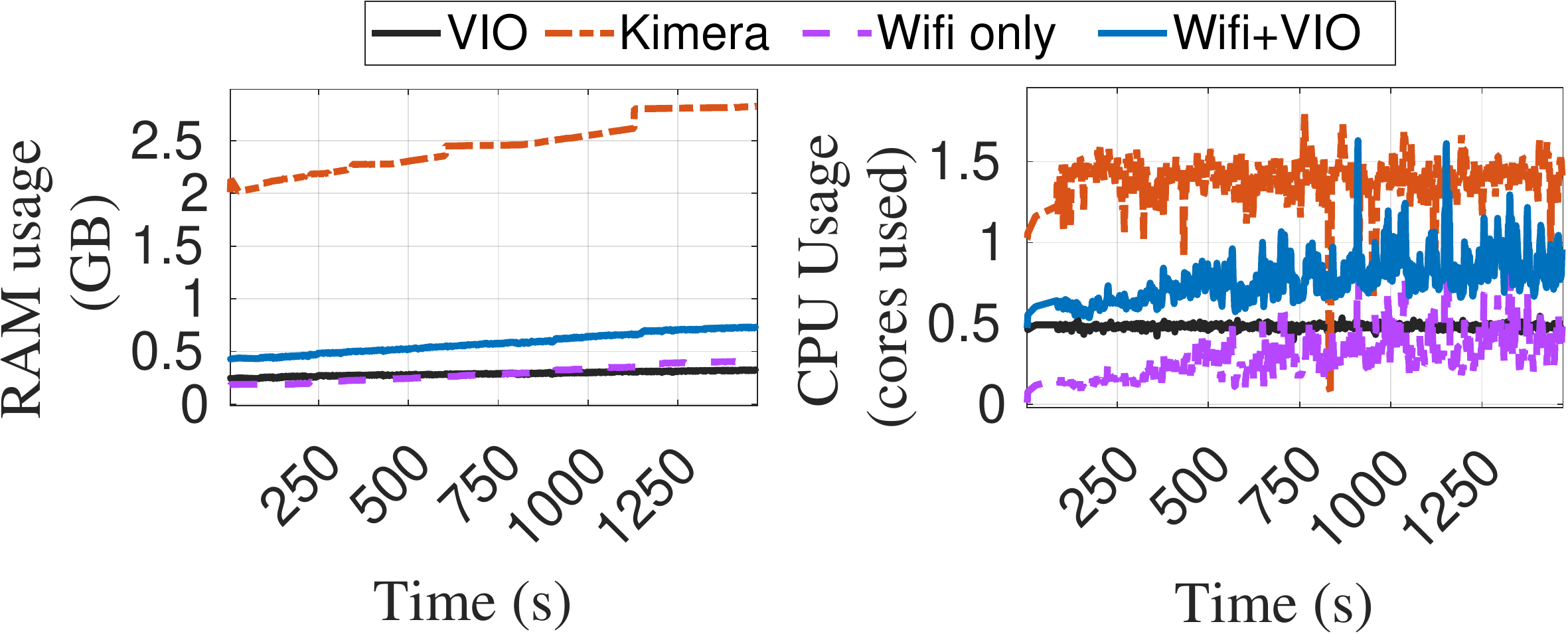}
    \end{minipage}
    \caption{Timeseries of memory (\textit{left}) and CPU consumption (\textit{right}) of VIO, Kimera and \name + VIO. The resource consumption of the \name's WiFi graph is also analysed (\textit{Wifi only})}
    \label{fig:mem-cpu}
\end{figure}


\noindent\textbf{Navigation Accuracy:} We compare the CDF and time-series translation errors in Figure~\ref{fig:end-end1}(a)(c) respectively for \name's WiFi+VIO, WiFi+Odom, and Kimera. 
From these plots we can see that the median ($90^{th}\%$) translation errors for \name's WiFi+VIO is $85$cm ($185$cm), and WiFi+Odom is $90$cm ($205$cm), which are $60\%$ lower compared to Kimera with median ($90^{th}\%$) translation errors of $105$cm ($300$cm).
Similarly Figure~\ref{fig:end-end1}(b)(d) compares the cumulative and time-series orientation error respectively for \name's WiFi+VIO, WiFi+Odom, and Kimera. 
From these plots we can see that the median ($90^{th}\%$) orientation errors for \name's WiFi+VIO is $3^{\circ}$ ($8^{\circ}$), and WiFi+Odom is $1^{\circ}$ ($4^{\circ}$), i.e. $60\%$ lower compared to Kimera with median ($90^{th}\%$) orientation errors of $5^{\circ}$ ($20^{\circ}$).

\begin{table*}[ht]
    \centering
    \footnotesize
    \caption{Translation, Orientation, and resource analysis of Cartographer and \name + LIO, median ($99^\mathrm{th} percentile$ error)}
    \label{tab:cart}
    \begin{tabular}{|l||cccc||cccc|}
    \hline
    \multirow{2}{*}{}                   & \multicolumn{4}{c||}{\textbf{Cartographer~\cite{cartographer}}}                                                                                                                    & \multicolumn{4}{c|}{\textbf{\name + LIO}}                                                                                                   \\ \cline{2-9} 
                                        & \multicolumn{1}{c|}{\textbf{\name DS}} & \multicolumn{1}{c|}{\textbf{DS 1}}       & \multicolumn{1}{c|}{\textbf{DS 2}} & \textbf{DS 3}        & \multicolumn{1}{c|}{\textbf{\name DS}} & \multicolumn{1}{c|}{\textbf{DS 1}}        & \multicolumn{1}{c|}{\textbf{DS 2}}         & \textbf{DS 3}      \\ \hline
    \textbf{Translation Error (cm)}     & \multicolumn{1}{c|}{\textbf{47.0 (98.9)}}            & \multicolumn{1}{c|}{74.0 (224)}          & \multicolumn{1}{c|}{134.1 (1097)}  & \textbf{23.3 (37.4)} & \multicolumn{1}{c|}{50.34 (152.4)}                     & \multicolumn{1}{c|}{\textbf{65.9 (92.2)}} & \multicolumn{1}{c|}{\textbf{67.3 (182.6)}} & 48.6 (114.4)       \\ \hline
    \textbf{Orientaion Error ($^\circ$)} & \multicolumn{1}{c|}{4.8 (7.9)}                        & \multicolumn{1}{c|}{\textbf{0.8 (4.66)}} & \multicolumn{1}{c|}{5.3 (12.6)}    & \textbf{0.6 (1.4)}   & \multicolumn{1}{c|}{\textbf{3.5 (6.9)}}               & \multicolumn{1}{c|}{2.4 (4.66)}           & \multicolumn{1}{c|}{\textbf{2.8 (12)}}     & 1.4 (3.3)          \\ \hline
    \textbf{Total Memory (MB)}          & \multicolumn{1}{c|}{520}                              & \multicolumn{1}{c|}{702}                 & \multicolumn{1}{c|}{\textbf{658}}  & 423                  & \multicolumn{1}{c|}{\textbf{486}}                     & \multicolumn{1}{c|}{\textbf{613}}         & \multicolumn{1}{c|}{706}                   & \textbf{356}       \\ \hline
    \textbf{Rate of Memory (MBps)}      & \multicolumn{1}{c|}{\textbf{0.30}}                    & \multicolumn{1}{c|}{0.29}                & \multicolumn{1}{c|}{0.33}          & 0.38                 & \multicolumn{1}{c|}{0.32}                             & \multicolumn{1}{c|}{\textbf{0.22}}        & \multicolumn{1}{c|}{\textbf{0.33}}         & \textbf{0.26}      \\ \hline
    \textbf{CPU (fraction of cores)}    & \multicolumn{1}{c|}{3.8 (4.4)}                        & \multicolumn{1}{c|}{3.2 (4.2)}           & \multicolumn{1}{c|}{3.8 (4.2)}     & 1.85 (4.4)           & \multicolumn{1}{c|}{\textbf{0.73 (1.90)}}             & \multicolumn{1}{c|}{\textbf{0.62 (2.1)}}  & \multicolumn{1}{c|}{\textbf{0.85 (2.8)}}   & \textbf{0.7 (1.1)} \\ \hline
    \end{tabular}
\vspace{-0.2in}
\end{table*}

While the median translation and orientation errors for Kimera are comparable to \name running on Kimera's odometry, the reason for high errors in the $90^{th}\%$ are due to an incorrect loop-closure that occurs in Kimera at around $1100$~sec time-mark as can be clearly seen from the sudden spike in errors in both Figures~\ref{fig:end-end1}(c,d). This can be seen in the two top-down views of the estimated trajectories shown in Figure~\ref{fig:end-end2}.
This further demonstrates the strength of WiFi-measurements for global corrections.

\noindent\textbf{Memory Consumption:} We have seen that WiFi provides more accurate and real-time global drift corrections than Kimera's loop closures. Now, we evaluate the memory consumption of the individual components of Kimera and \name and observe the trends shown in Figure~\ref{fig:mem-cpu} (left), from which we can see that while Kimera needs a start up memory of $2$ GB and from then on accumulates memory at a rate of $0.56$ MBps. In contrast, WiFi + VIO only has a start-up memory of $0.4$ GB and memory accumulation rate of $0.2$ MBps, which will on an average enable \name to run $3.3\times$ longer than Kimera on a typical RAM of $8$ GB before crashing.
We can further verify that Kimera's memory consumption is dominated by loop closures -- the black line in Figure~\ref{fig:mem-cpu} (left) for the VIO, which is Kimera without loop-closures, indicates the memory remains constant at $0.2$ GB. It is important to note here that since Kimera's VIO has near-constant memory consumption, the accruing memory consumption of \name's stack is purely due to the WiFi-graph as shown by the purple dashed line in Figure~\ref{fig:mem-cpu} (left).
We note that this could be avoided if the WiFi-graph employs a fixed-lag smoothing optimization strategy similar to the VIO system, which will be left for future work.

\noindent\textbf{Compute Requirements:} Finally, we can also see from Figure~\ref{fig:mem-cpu} (right) that \name requires only one core to run on our system, while Kimera takes up to $1.5$ cores owing to its loop closure detection and correction algorithms. Thus \name's WiFi+VIO design requires about $3.3\times$ less memory and $1.5\times$ less compute than a state-of-the-art Kimera system while achieving about $60\%$ better navigation performance.

\begin{figure*}[pt]
    \begin{minipage}[t]{0.24\textwidth}
        \includegraphics[width=\textwidth]{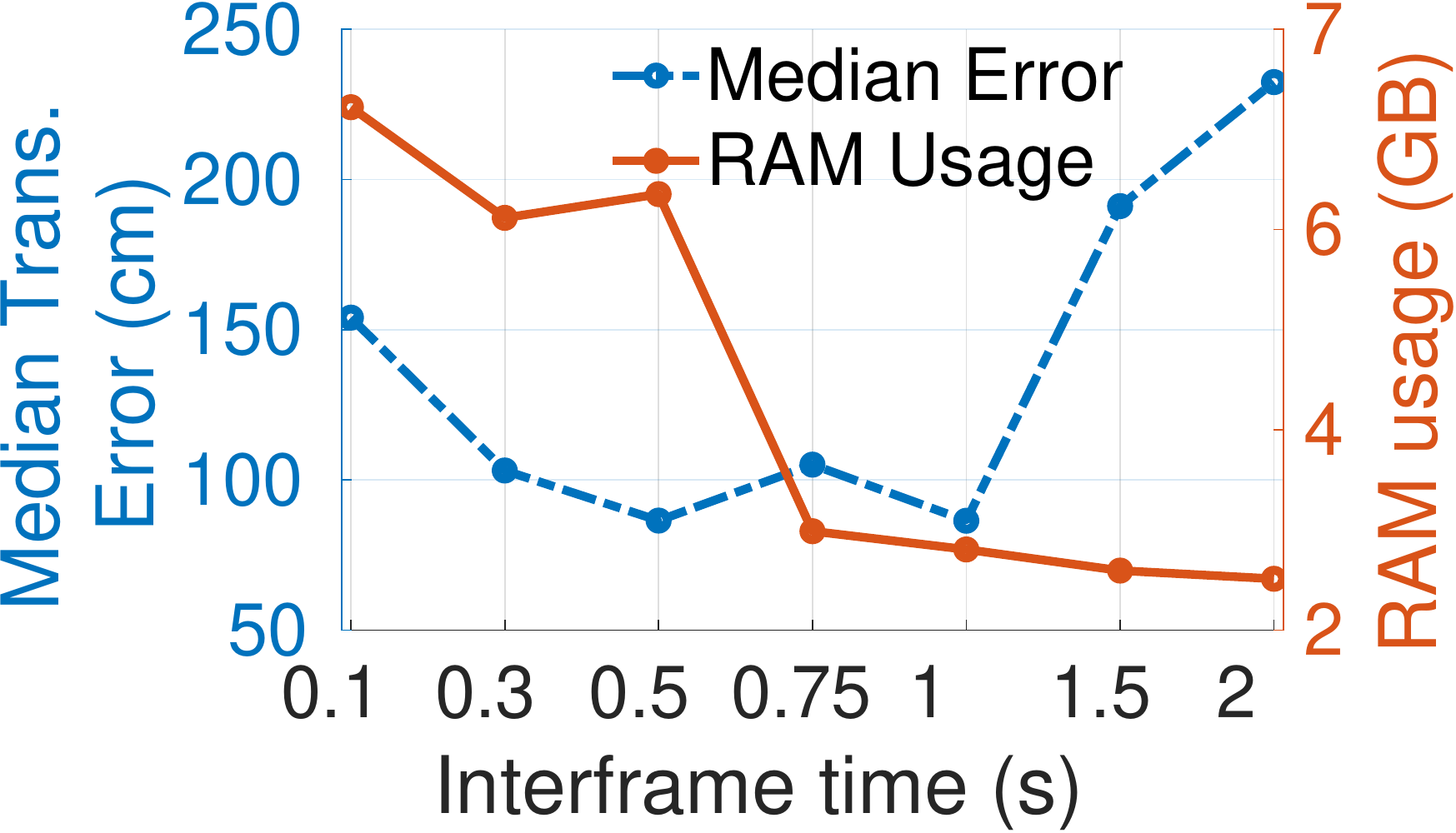}
        \subcaption{}
    \end{minipage}
    \begin{minipage}[t]{0.24\textwidth}
        \includegraphics[width=\textwidth]{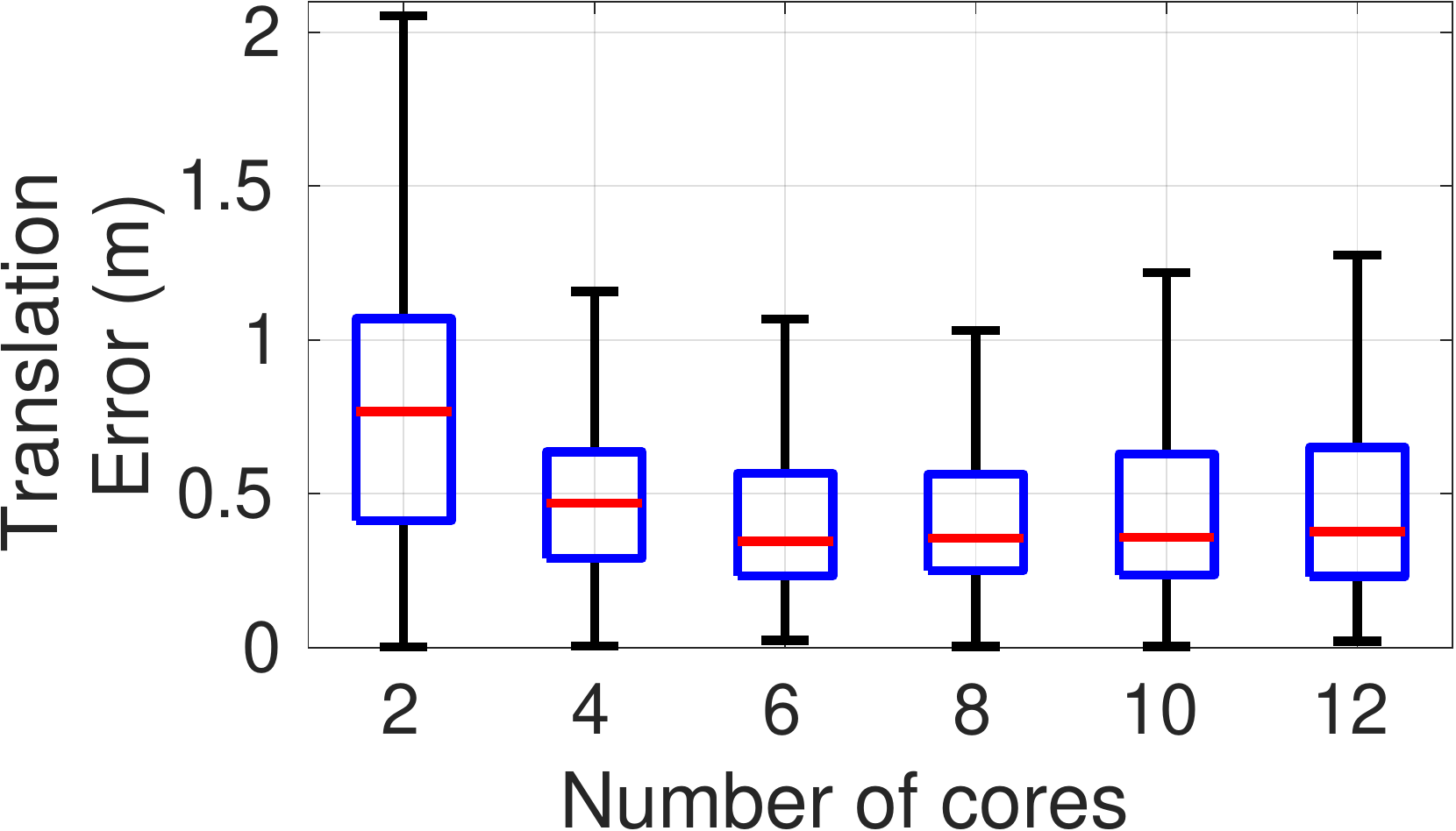}
        \subcaption{}
    \end{minipage}
    \begin{minipage}[t]{0.24\textwidth}
        \includegraphics[width=\textwidth]{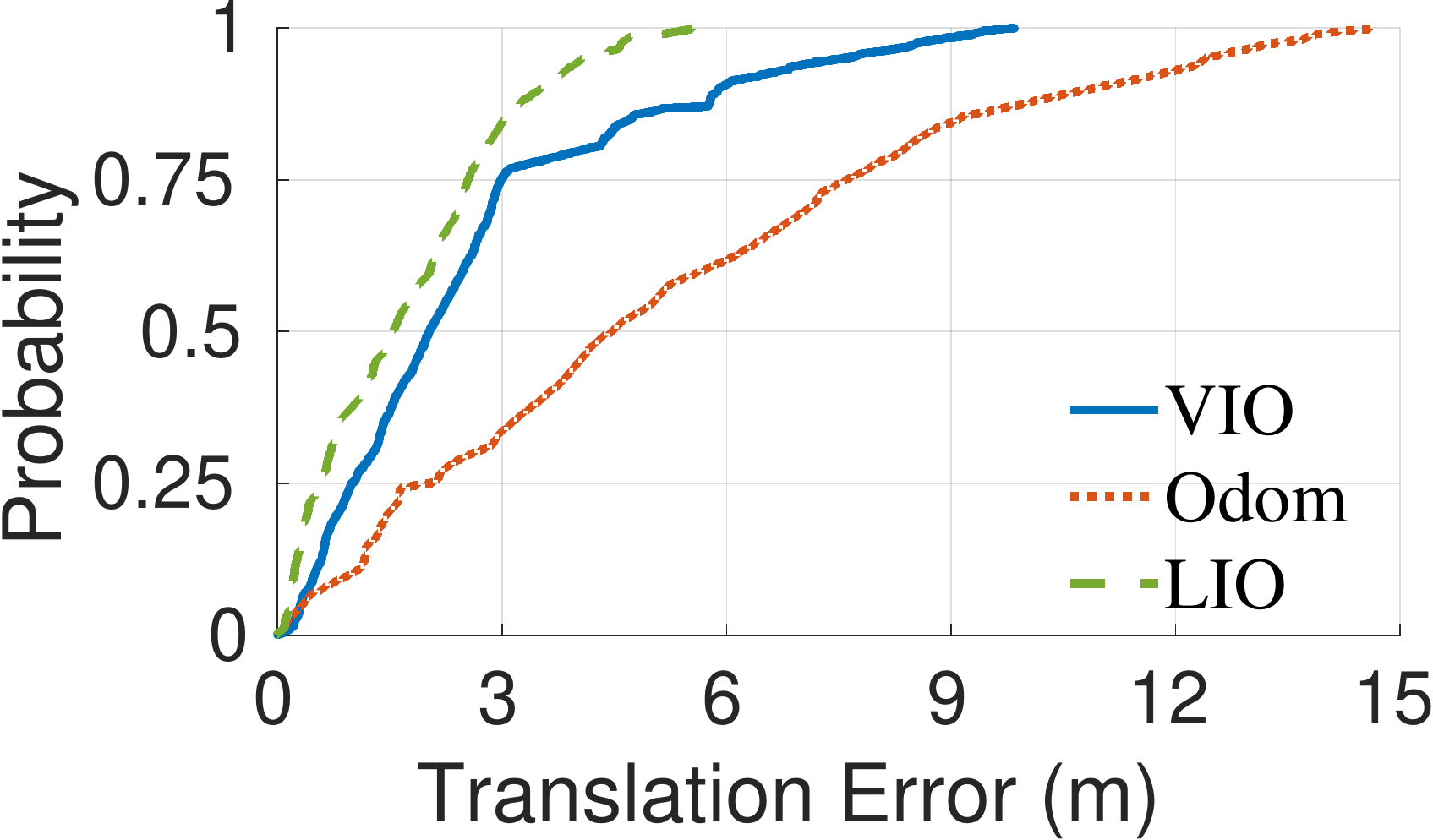}
        \subcaption{}
    \end{minipage}
    \begin{minipage}[t]{0.24\textwidth}
        \includegraphics[width=\textwidth]{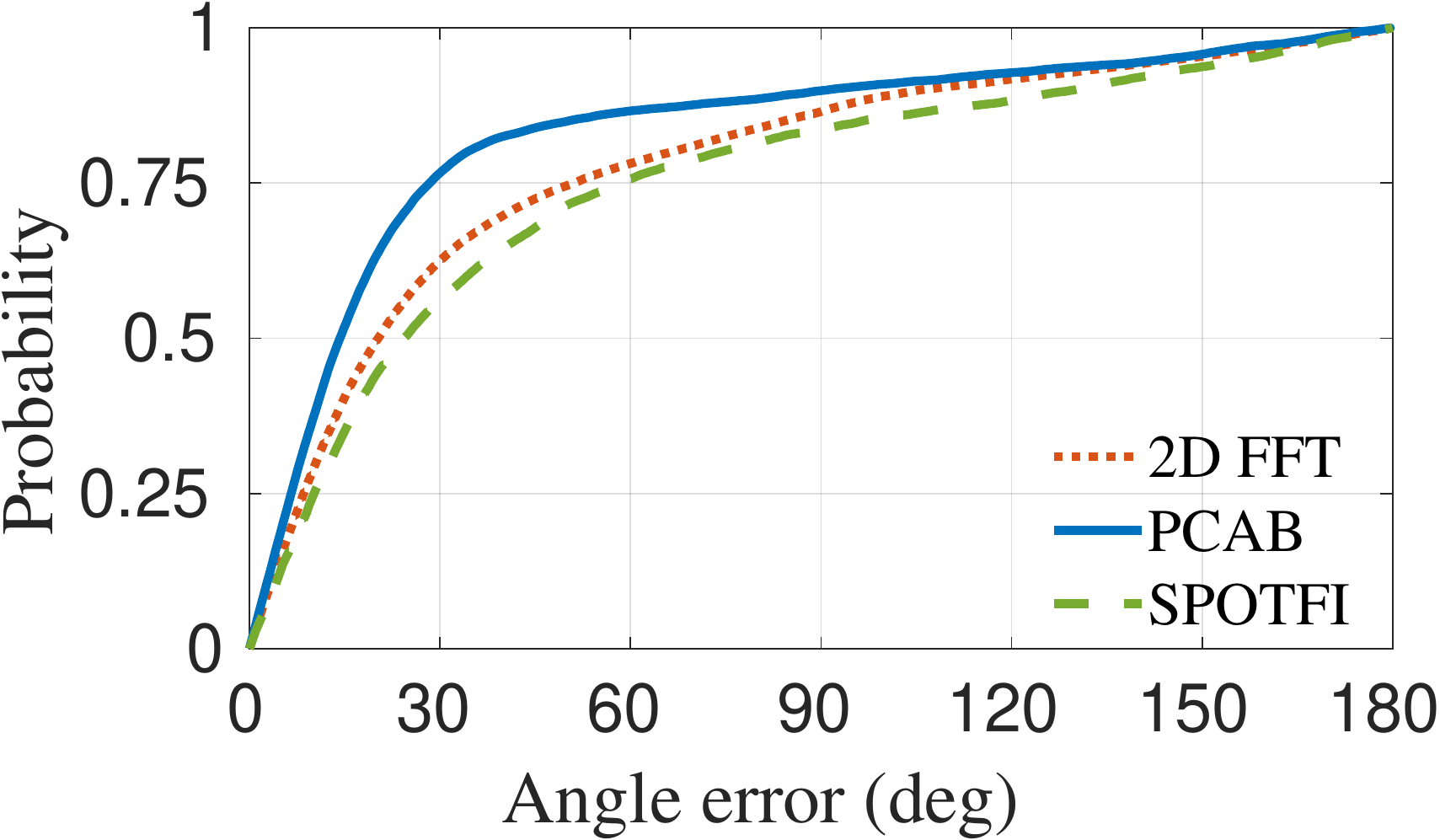}
        \subcaption{}
    \end{minipage}
    \caption{\textbf{Microbenchmarks}: \textbf{(a) Kimera:} Trade-off between memory consumption and the accuracy for various interframe rates. \textbf{(b) Cartographer:} Trade-off between compute required and the accuracy for various scan-matching thresholds. \textbf{(c) Odometry-Graph:} Comparison on how various odometry inputs to the WiFi graph look before optimization. \textbf{(d) Bearing Measurement Algorithms:} Comparison between 2D-FFT from $\mathrm{P^2}$SLAM, Spotfi and our proposed PCAB.}
    \label{fig:micro}
    \vspace{-0.1in}
\end{figure*}

\subsection{End-to-end evaluation (\name with Cartographer):}\label{sec:eval-cart}

We also test \name running on Cartographer's LIO and compare it online with full-stack Cartographer across all the 4 datasets, DS 1/2/3 and \name DS and present a summary of the results in Table~\ref{tab:cart} due to space constraints.

\noindent\textbf{Navigation Accuracy:} From this table, we can see that \name's trajectory estimation performs on-par with Cartographer's loop closure detector across most datasets in terms of translation and orientation accuracy. There are two notable differences in navigation accuracy performance: 

\textbf{(a)} in \textit{DS 2} \name's median ($90^{th}$) translation and orientation errors are $2\times$ ($5\times$) lower for \name + LIO than Cartographer. In this scenario, Cartographer makes an incorrect loop closure leading to a very inconsistent trajectory prediction, as can be common with visual-based loop closures. 

\textbf{(b)} in  \textit{DS 3}, Cartographer has $2\times$ lesser translation and orientation errors than WiFi+LIO. We notice this performance degradation from the use of linear array on the robot. As discussed in \cref{sec:design-bearing}, bearing measurements closer to $\pm 90^\circ$ suffer from higher errors, and due to a non-optimal orientation of linear array, we observe a larger number of bearing measurements at these higher angles, reducing the SLAM performance. Utilizing a square array resolves these issues. 

\noindent\textbf{Memory Consumption:} The memory consumption of \name and Cartographer are similar. This occurs because, unlike Kimera's VIO system where each camera frame consists of dense features, Cartographer's LIO system's LiDAR scans are much smaller, reducing Cartographer's memory consumption.

\noindent\textbf{Compute Requirements:} Due to the sparsity of LiDAR features, Cartographer needs to run scan-matching algorithms for loop closures, which in turn increases Cartographer's compute requirements. In contrast, since \name running on Cartographer's local odometry does not require scan matching to correct the global trajectory, it demands much lesser compute resources. This can be observed in the last row of Table~\ref{tab:cart} which shows the average number of cores used over the entire run of the robot navigation with the maximum number of cores used in parentheses. We can see that \name needs $4\times$ lesser peak compute than the full-stack Cartographer implementation across all 4 datasets.

\subsection{Microbenchmarks}\label{sec:eval-micro}

\noindent{\textbf{VIO Memory vs Accuracy:}}
To characterize Kimera's memory usage, we alter the rate at which it records keyframes for loop closure detection. A lower time between keyframes will lead to more loop closure detections at the cost of increased memory consumption. To understand how Kimera's loop closure detection accuracy is limited by memory, we plot the median translation error and the total memory consumed in \name DS (Figure~\ref{fig:micro}(a)). From this plot, we can see that the best median translation error with reasonably low memory consumption occurs at a keyframe period of $1$~second and we use this keyframe rate for our baseline.

\noindent{\textbf{LIO Compute vs Accuracy:}}
To next understand the performance of Cartographer we first note that single-plane-based LIO systems have sparse features in LiDAR scans, as opposed to denser stereo data. This sparser representation demands extensive scan-matching algorithms~\cite{cartographer} with higher compute. Thus to understand Cartographer's compute requirements, we run Cartographer on \name DS and limit the number of CPU cores it can access. We plot the median translation accuracy vs number of CPU cores accessible in Figure~\ref{fig:micro}(b). We can see that median translation error improves with a larger number of cores, verifying that compute power is a bottleneck for accurate scan matching, and thus navigation performance.
Keeping in mind that many low-compute platforms are limited to 4 cores, we restrict Cartographer to use 4 compute cores only in the end-to-end comparison.

\noindent{\textbf{Wheel-based Odometry vs Kimera without LCD (VIO) and Cartographer without LCD (LIO):}}
Now, let us compare how accurate the odometry measurements from Kimera and Cartographer without loop closures are to wheel odometry provided by the Turtlebot.
Figure~\ref{fig:micro}(c) shows the comparison of their translation errors, from which we can see that the odometry measurements from Kimera and Cartographer without loop closure have lesser errors at both the median and $90^{th}\%$ than odometry from wheel-encoders.
This supports the dual-graph design intuition of \name wherein the odometry from the VIO/LIO-graph is locally corrected and so is better at local odometry and mapping than simple wheel-encoder systems.

\noindent{\textbf{Bearing Accuracy and Compute:}}
Finally, to understand the accuracy and compute-requirements of bearing-estimation algorithms and the need PCAB (\cref{sec:design-bearing}), we compare it with \textit{2D-FFT} defined in $\mathrm{P^2}$SLAM~\cite{arun2022p2slam} and \textit{Spotfi}~\cite{spotfi}. As shown in Figure~\ref{fig:micro}(d), PCAB outperforms 2D-FFT in bearing estimation by $1.43\times$ ($2 \times$) at the median and $80^\mathrm{th}\%$. Additionally, we note that the average CPU utilization of PCAB on our robot is $11\%$, whereas 2D-FFT required $565\%$ core usage (an improvement of approx. $50\times$). Moreover, we compute the AoA predictions with Spotfi~\cite{spotfi} and find PCAB performs $1.8\times$ ($2.2\times$) at the median and $80^\mathrm{th}\%$. Moreover, Spotfi consumes $1200$ percentage of cores (all cores of our machine). Clearly, both the 2D-FFT and Spotfi estimations are unsuitable for low-compute SLAM applications.
\section{Limitations and Future Work}\label{sec:conclusion}



In this work, we have demonstrated \name and its modular design can integrate into any existing VIO/LIO systems, to remove the need of compute and memory intensive loop-closures. Thus, \name provides a framework that can enable accurate, real-time, and resource-efficient SLAM compared to Kimera~\cite{rosinol2021kimera} and Cartographer~\cite{cartographer} for indoor robotics. However, there are still some limitations of \name's novel framework that opens new avenues for future work:

\noindent\textbf{(a)} While \name demonstrates loop-closure free SLAM through WiFi, any RF-sensor that can measure bearings to unique landmarks in the environment would work. In particular, UWB localization systems~\cite{zhao2021uloc} can be easily deployed in our framework to furnish 6 DoF poses.

\noindent\textbf{(b)} We have seen how different antenna array geometry on the robot provides varied results, extending the work to perform single-antenna based WiFi-SAR algorithms as demonstrated in WSR~\cite{jadhav2020wsr} would make the system more scalable.

\noindent\textbf{(c)} While \name demonstrates efficient SLAM for a single robot in the environment, extensions to collaborative SLAM for a fleet of robots presents additional challenges, making compute and memory efficiency even more important.



\bibliographystyle{IEEEtran}
\bibliography{root}
\end{document}